# Generation of Chinese classical poetry based on pre-trained model

| First author | Second author | Third author |
|---|---|---|
| Ziyao Wang | Lujin Guan | Guanyu Liu |
| School of Information Science Engineering | College Of Life Science Technology Huazhong University Of Science And Technology | School of Computer and Cyberspace Security Communication University of China |
| Lanzhou University | | |
| 2126546982@qq.com | guanlujin@gmail.com | lyyuu168@163.com |

## Abstract

In order to test whether artificial intelligence can create qualified classical poetry like humans, the author proposes a study of Chinese classical poetry generation based on a pre-trained model. This paper mainly tries to use BART and other pre training models, proposes FS2TEXT and RR2TEXT to generate metrical poetry text and even specific style poetry text, and solves the problem that the user's writing intention gradually reduces the relevance of the generated poetry text.

In order to test the model's results, the authors selected ancient poets, by combining it with BART's poetic model work, developed a set of AI poetry Turing problems, it was reviewed by a group of poets and poetry writing researchers. There were more than 600 participants, and the final results showed that, high-level poetry lovers can't distinguish between AI activity and human activity, this indicates that the author's working methods are not significantly different from human activities. The model of poetry generation studied by the author generalizes works that cannot be distinguished from those of advanced scholars.

The number of modern Chinese poets has reached 5 million. However, many modern Chinese poets lack language ability and skills as a result of their childhood learning. However, many modern poets have no creative inspiration, and the author's model can help them. They can look at this model when they choose words and phrases and they can write works based on the poems they already have, and they can write their own poems. The importance of poetry lies in the author's thoughts and reflections. It doesn't matter how good AI poetry is. The only thing that matters is for people to see and inspire them.

关键词： Deep Learning；Text Generation；Pre-trained model；BART；Classical Chinese poetry；Turing test

## 1 Introduction

Chinese classical poetry occupies an important position in Chinese classical literature and has a far-reaching influence, among which Juju is a very important part, which is also called quatrain (Lai, 2020). It can be divided into five words and seven words, no matter which one needs to follow the structural, tonality and semantic requirements of ancient poetry. Word to word and line to line all meet specific phonological patterns. For example: The last character of the second, fourth and (possibly) first lines must rhyme, and the third line has no restrictions. In addition, poetry must follow a prescribed tone pattern: Tones or tones of the tones. In addition, Chinese classical poetry has always had a high requirement for semantic meaning, every ancient poem must convey the author's emotion through the verse, so that readers can feel the author's state of mind and inner feelings. Nowadays, with the progress of technology, it is possible to use machines to generate quatrains (Nenu, 2022). The research on automatically generated poetry began in the 1960s and has been a focus in recent decades. In the field of generating ancient poems, after experiencing the use of templates, genetic algorithms, based on other text generation methods. With the focus on the intersection of deep learning and natural language process, neural networks have been widely used in poetry creation (Ma, 2021).

In 2020, the number of people who can be counted to create poetry in China has reached 5 million (Dayong and Yufei, 2021). However, due to the great difference between the language used in classical poetry and modern Chinese, most creators often encounter language difficulties when creating poetry, and can't find suitable words to express their feelings. In order to help millions of today's poetry creators create better, it is meaningful to explore how to use the depth learning model to generate modern poetry.

The main contributions of this paper are:

- We have constructed the most complete data set of Chinese classical poetry.

- This paper uses pre-trained models such as BART on this dataset, and proposes two different poetry generation models for different scenarios. The generation of item classical poetry will encounter the problem that the relevance between the sentences behind the generated works and the user's writing intention gradually decreases.

- This paper proposes related algorithms to generate poems of different styles.

- We conducted a Turing like test, and found that the poetry lovers with high level could not distinguish the works generated by the model trained in this paper from the works of human beings. Discusses the limitations of AI and related ethical issues.

## 2 Related research

As a long-term concern of artificial intelligence, the research of poetry automatic generation can be traced back to decades ago. The first step in this field is based on rules and templates (Gervás, 2001). Since the 1990s, statistical machine learning methods have been used to generate poetry, such as genetic algorithm (GA) (Manurung, 2004)and statistical machine translation (SMT) (He et al., 2012).

After the rise of in-depth learning, great advantages have been achieved in the generation of poetry text. The existing attempts to generate Chinese metrical poetry text are relatively successful in the Nine Songs AI (THUNLP, ) and the Three Hundred Poems AI (wangjiezju1988, ) of Tsinghua University. Jiuge was developed by the Natural Language Processing and Social Humanistic Computing Laboratory of Tsinghua University. It mainly uses GRU algorithm and Sequence to Sequence model (Yi et al., 2018), which is trained on the corpus of more than 300000 metrical poems, and has achieved good results. Shi300 goes further, using the Chinese BERT model opened by Harbin Institute of Technology as a pre training (Cui et al., 2021), further expanding the corpus to more than 800000 works (Werneror, 2018), and using the Sequence to Sequence model to generate topic to text. Shi300 is the best one among the well-known poetry generation websites at present. In order to strengthen the correlation between the sentences behind the generated works and the user's writing intention, Jiuge retains the most significant part of each poem, and then generates the next poem through the theme and the saved information of the previous sentence 11 et al. divided poetry generation into two stages: what to say and how to say. First, generate several sub topics based on the user's input writing intention, and then generate each poem based on the sub topics. These studies have achieved some results. As for the generation of poems with specific styles, (Liu et al., 2018) designed a reward function for style matching by calculating the cosine distance between the generated poems and poems with specific styles, which can generate high-quality poems with three specific styles 13 et al. separate poems of different styles through mutual information, and generate output of specific styles according to manually selected style input. With the development of natural language processing technology, the pre training model has been applied more and more in ancient Chinese literature. Zhao Zhe et al. (Zhao et al., 2019) trained the gpt2 chinese poem model with the data of about 800000 poems to continue writing poems.Wang Dongbo et al. (Dongbo et al., 2021) trained SikuBERT with the four libraries of complete books as the language material to complete the tasks of punctuation of ancient Chinese sentences and named entity recognition.The BART model (Lewis et al., 2019) selected in this paper was proposed by Facebook. Compared with BERT and other pre training models, the pre training language model using the overall structure of Transformer model has no decline in the performance of natural language understanding tasks, and has significantly improved in natural language generation tasks. On the basis of these studies, this study first trained a general BART poetry model to complete the seq2seq task, strengthened the relevance between the whole poem and the user's writing intention by specifying the theme words and keywords in the poem, and generated poems of specific style by controlling the theme words and keywords, which achieved good results.

## 3 Method

### 3.1 BART-poem

Although there are already pre training models applicable to modern Chinese or ancient Chinese, the language of classical poetry is quite different from modern Chinese, and the combination of words and the linking logic of words and sentences are also different from ancient Chinese in terms of pragmatics and grammar. Therefore, it is necessary to train a pre training model suitable for poetry tasks. By evaluating the effects of different models, this paper selects the BART model from the BERT, Roberta, T5 and BART models for training. The size of this model is about 1.5GB. The main parameters are as follows:

| embedding | feedforward | hidden | heads | layers | dropout | encoder&decoder |
|---|---|---|---|---|---|---|
| 1024 | 4096 | 1024 | 16 | 12 | 0.1 | transformer |

Table 1: BART模型参数

This article uses the open source UER of the target (Zhao et al., 2019) The py item is used for model training. First, add the Chinese characters whose number of occurrences in the poetry corpus is more than or equal to 100 to the vocabulary, set the sequence length to 64, and specify the data processor as the bart mode to preprocess the data. Then set the batch size to 64, span max length to 3, and train 60000 steps. Finally, the accuracy of the model is stable at 0.91, and the loss is about 0.50. We name this model BART poem.

### 3.2 Data process

#### 3.2.1 Theme words extraction

First, the poetry text is segmented with THULAC (Yang et al., 2018) developed by Tsinghua University, then the words in the stop list are removed, and the remaining words are extracted with TF-IDF algorithm. TF-IDF is a statistical method to evaluate the importance of a word to a file set or one of the files in a corpus. The importance of a word increases with the number of times it appears in the file, but decreases with the frequency of its appearance in the corpus. In this paper, the algorithm is used to extract subject words. The number of text subject words extracted from each poem is 1/12 of the length of the poem text.

#### 3.2.2 Key chars extraction

This paper has noticed the concept of "eye of poetry" in the field of classical poetry. If a word in poetry is the center of the context and the meaning of other words revolves around this word, then this word is the eye of poetry, which is regarded as a key word in this paper.

In this paper, we choose to use the word vector (Li et al., 2018) trained by Shen Li et al. on the corpus of the four libraries of the whole book using word2vec algorithm. After the stop words are removed from the poetry text, the remaining words are converted into word vectors for representation. Through calculation, we determine the center point of these word vectors and find several vectors closest to the Euclidean distance from the center point. The characters corresponding to these vectors are the keywords of poetry. The number of keywords extracted from each poem is 1/10 of the length of the poem text.

### 3.3 FS2TEXT

#### 3.3.1 Overview

Many poetry generation models in the past regarded the task of poetry generation as "the mapping from title to text". However, there is no clear correspondence between the title of poetry and the content of poetry, such as the famous "falling flowers poetry". At present, there are at least 1000 poems with the title of "falling flowers". Although the title is the same, their content is different, so there is no appropriate mapping relationship between the title of poetry and the poetry text. So this paper decided to change the main mapping relationship into a one-to-one mapping "from the first sentence to the whole poem". In order to solve the problem that the relevance between the sentences behind the works and the users' writing intentions gradually decreases, this paper uses theme words and keywords to control the generation process of poetry and the generation of specific style works.

#### 3.3.2 FS2TEXT model structure

Execute the seq2seq task on the basis of the trained BART poem model. The input is "the first sentence of each poem &the theme words of the random number of this poem &the keywords of the random number of this poem&the poem type", and the output is the full text of the poem.

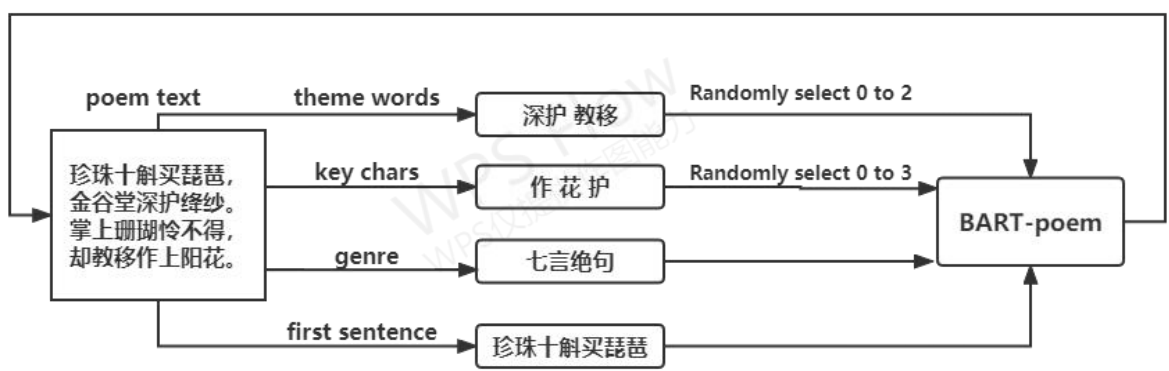

Figure 1: FS2TEXT

#### 3.3.3 Generation of speciftc style poetry

If you want to generate poems with a specific style, you need to construct a data set that only contains poems with a specific style, extract all the subject words and keywords, and use the subject words and key words in the data set that only contains poems with a specific style on the basis of the trained FS2TEXT model, input them into the model together with the genre and the first sentence to get a poem with a specific style.

### 3.4 RR2TEXT
#### 3.4.1 Overview
The infra rhyme is a kind of creation behavior using the same genre and rhyme as the target works, which is very popular in poetry creation. Du Fu's Eight Poems of Autumn Prosperity, Wang Shizhen's Four Chapters of Autumn Willows, Huang Jingren's Sixteen Poems of Qihuai and other poems have been infra rhymed repeatedly by later generations. We can abstract the action of infra rhyme as: creating works that are similar to the original style and have the same rhyme and style. The first sentence of the second rhyme work should not be the same as the original work, so this paper uses rhymes to generate the second rhyme poetry work, and uses the keywords and subject words of the original poem to control the infra rhyme work, which is similar to the original style.

#### 3.4.2 RR2TEXT model structure
Execute the seq2seq task on the basis of the trained BART poem model. The input is "the rhyme of each poem&the theme word of the random number of this poem&the keyword of the random number of this poem", and the output is the full text of the poem. Because the style and content of the work of infra rhyme are similar to the original work, we can control the generated work by inputting the same subject words and keywords as the original work, and most of the work of infra rhyme will not be very similar to the content of the original work, so this paper will only input some subject words and keywords of the original work to ensure that the generated work will not be highly similar to the original work when performing the generation task.

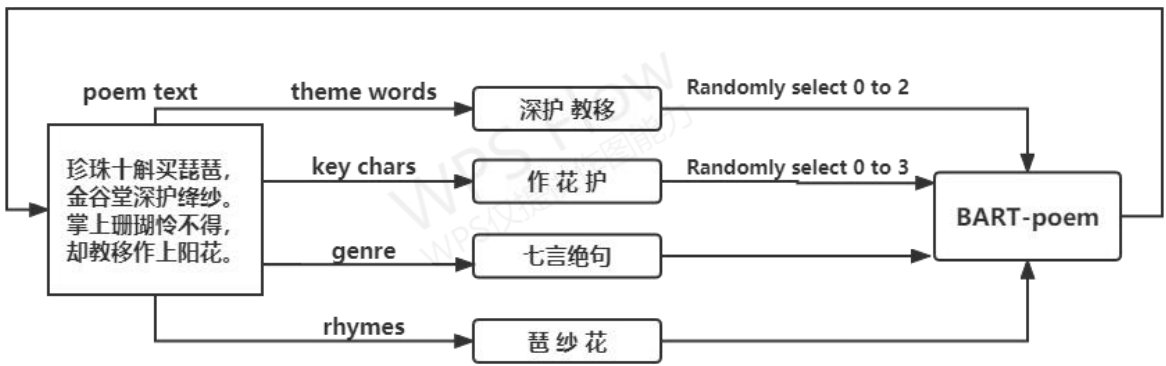

Figure 2: RR2TEXT

## 4 Experiments
### 4.1 Dataset
This project has constructed the most complete poetry data set currently open, and put poetry works of all ages in a csv file, with a total of about 1.2 million pieces, which are stored in four fields: "title", "dynasty", "author", and "content".

| bf Title | bf Times | bf Author | bf Poetry Text |
|---|---|---|---|
| 失题 | 当代 | 杜随 | 后会何须约，前尘自可忘。一时同梦寐，万古各参商。 |

Table 2: Data Storage Example

The open source project of Werneror 10, the user of item GitHub, has collected about 800000 ancient poems from pre Qin to modern times. The ancient poetry data is stored in four fields: "title", "dynasty", "author", and "content".

- The open source project of Werneror 10, the user of item GitHub, has collected about 800000 ancient poems from pre-Qin to modern times. The ancient poetry data is stored in multiple csv files by dynasty, with four fields: "title", "dynasty", "author" and "content

| bf Title | bf Times | bf Author | bf Poetry Text |
|---|---|---|---|
| 失题 | 当代 | 杜随 | 后会何须约，前尘自可忘。一时同梦寐，万古各参商。 |

Table 1: Data Storage Example

There are some rare words in the poetry corpus that cannot be displayed, so "?" is used instead.

- The item Werner 10 project lacks many works of poets in the Ming and Qing Dynasties. This paper collects many works of poets in Jin, Yuan, Ming and Qing Dynasties from various poetry websites through web crawlers.

- Because of some rare poetry materials, item has no public digital resources on the internet. In this paper, PDF files of ancient books are manually entered to make them easy to handle.

### 4.2 FS2TEXT training

On the dataset constructed in this paper, use the BART poem model to fine tune and execute the seq2seq task. Set the sequence length to 64 and specify the data processor as the bart mode to preprocess the data. Then set the batch size to 64 and the training loss to about 2.60. At this time, through manual judgment, the model has been able to generate poems of acceptable quality, stop training, and get the FS2TEXT model. At this time, specify the genre, input the first sentence and a certain number of keywords and subject words, and the corresponding poetry can be generated. The user's intention can affect the generated text through keywords and subject words.

| Input | Output |
|---|---|
| 七言绝句 | 杨柳花飞芜草青 |
| 白鹭 | 野塘烟草自凋零 |
| 烟—山 | 一双白鹭来烟际 |
| 杨柳花飞芜草青 | 点破遥山数抹青 |

Table 2: FS2TEXT example

The results generated by the model conform to the rules and formats of the specified genre, and it can be seen that subject words and keywords affect the overall text.

### 4.3 FS2TEXT-amorous

This paper selects some works of Wang Yanhong and Sun Yuanxiang as the data set of gaudy poems. After extracting the subject words and keywords in the gaudy poems data set, the trained FS2TEXT model specifies that only the subject words and keywords in this data set can be used to generate gaudy poems.

| Input | Output |
|---|---|
| 七言律诗 | 相见时难别亦难，临歧无奈暂盘桓。 |
| — | 舟沿碧草同千里，人隔青天共一峦。 |
| — | 梦去不妨风浩荡，酒来犹喜月团圆。 |
| 相见时难别亦难 | 从今珍重琼瑶字，莫作鸳鸯万缕看。 |

Table 3: FS2TEXT-amorous example

### 4.4 RR2TEXT training

Fine tune the BART poem model and execute the seq2seq task. Set the sequence length to 64. Specify the data processor as the bart mode to preprocess the data. Then set the batch size to 64 and the training loss to about 2.80. At this time, through manual judgment, the model has been able to generate poems of acceptable quality. Stop training and get the RR2TEXT model. The user inputs the works that want to infra rhyme into the program. The program analyzes the genre, rhyme, subject words, keywords, etc. after processing, and inputs them into the model to get infra rhyme works.

| Input | Output |
|---|---|
| 独起凭栏对晓风 | 日没荒墟生晓风 |
| 满溪春水小桥东 | 满溪流水碧山东 |
| 始知昨夜红楼梦 | 不知渔父相扶醉 |
| 身在桃花万树中 | 独立苍茫烟雨中 |

Table 4: RR2TEXT example

### 4.5 test

#### 4.5.1 test rules

In order to verify the quality of the poetry works generated by the model, this paper imitates Turing test to produce a set of test questions. The process is as follows:

- selects important poets in literary history (representing a higher level of mankind)

- selects some unknown works from the works of these poets (to avoid the influence on the experimental results caused by someone reading them)

- According to the genre, item selects 4 pieces from each genre, totaling 16 pieces.

- After the item is selected, enter the first sentence to make the FS2TEXT model generate a work. Put the work generated by the FS2TEXT model together with the original work to choose one from the other, so that people can judge which song is created by AI.

In the end, 616 valid answers were received. The test participants included professional researchers from famous universities such as Peking University and Fudan University. On the whole, the subject's accomplishments in poetry were significantly higher than those of ordinary people.

#### 4.5.2 Result analysis

According to the questionnaire, the accuracy of most questions fluctuated around 50%, and only three questions were below 40% or above 60%. In general, the testers made 9856 choices, 4960 of which were correct, accounting for 50.32% of the overall proportion, which is very close to 50%. Therefore, we concluded that even for those who are more professional and have higher poetic attainments, it is difficult to distinguish AI's works.

## 5 Conclusion and discussion

During the Turing like test, some poetry creators saw the level of AI's poetry generation and expressed concern that someone might take AI's poetry to participate in the poetry contest. This paper also has no good solution to this problem. The only thing we can do is to appeal to Jiuge, Shi300, a poetry generation website, to open the historical works generated by its model, so that we can judge whether a poem is written by ourselves or generated by the poetry generation model of these websites through query.

The technology in the field of natural language processing is changing with each passing day, and the quality of poetry generated by AI is also rising. The poetry generation model studied in this paper has been able to write works that cannot be distinguished by high-level researchers. The number of contemporary Chinese poetry creators has reached 5 million. However, due to the education they received from childhood, a considerable number of contemporary creators lack language sense and skills. The model in this paper can help them. When they choose words and sentences, they can refer to this model to generate works based on their existing poems to create their own poems. The significance of poetry lies in the author's thoughts and feelings. No matter how well AI's poetry is written, it is meaningless - its only significance is to let people see and inspire.